\documentclass[preprint,12pt]{elsarticle}

\usepackage{xcolor}
\usepackage{enumitem}
\usepackage{amsmath,amssymb}
\usepackage{multirow}
\usepackage{url}  
\makeatletter
\g@addto@macro{\UrlBreaks}{\UrlOrds} 
\makeatother

\begin{document}

\begin{frontmatter}



\title{Human-Assisted Robotic Detection of Foreign Object Debris Inside Confined Spaces of Marine Vessels Using Probabilistic Mapping}

\author[1]{Benjamin Wong}
\author[1]{Wade Marquette}
\author[1]{Nikolay Bykov}
\author[2]{Tyler M. Paine}
\author[1,3]{Ashis G. Banerjee \corref{cor1}}

\address[1]{Department of Mechanical Engineering, University of Washington, Seattle, WA 98195, USA}
\address[2]{Naval Undersea Warfare Center Division Keyport, Keyport, WA 98345, USA}
\address[3]{Department of Industrial \& Systems Engineering, University of Washington, Seattle, WA 98195, USA}

\cortext[cor1]{Corresponding author; Email address: ashisb@uw.edu}

\begin{abstract}
Many complex vehicular systems, such as large marine vessels, contain confined spaces like water tanks, which are critical for the safe functioning of the vehicles. It is particularly hazardous for humans to inspect such spaces due to limited accessibility, poor visibility, and unstructured configuration. While robots provide a viable alternative, they encounter the same 1q8set of challenges in realizing robust autonomy. In this work, we specifically address the problem of detecting foreign object debris (FODs) left inside the confined spaces using a visual mapping-based system that relies on Mahalanobis distance-driven comparisons between the nominal and online maps for local outlier identification. The identified outliers, corresponding to candidate FODs, are used to generate waypoints that are fed to a mobile ground robot to take camera photos. The photos are subsequently labeled by humans for final identification of the presence and types of FODs, leading to high detection accuracy while mitigating the effect of recall-precision tradeoff. Preliminary simulation studies, followed by extensive physical trials on a prototype tank, demonstrate the capability and potential of our FOD detection system. 
\end{abstract}




\end{frontmatter}


\section{Introduction}
\label{sec:intro}

Most large marine vessels are complex systems that operate in extreme open-ocean environments.  As a result, they require significant cost and effort to maintain (inspect and repair), especially as the vessels age.  For example, the total maintenance cost of the US Navy vessels was approximately 20 billion US dollars in 2020 \cite{navy}, and this cost is projected to increase as the size of the fleet increases.  Some of the most challenging maintenance tasks occur inside large tanks and other confined spaces inside the vessels. In particular, the vessels contain numerous ``grey-water'' tanks that can be fully or partially filled with seawater when the vessels are underway. The tanks provide critical access to much of the machinery on the vessels, but are difficult and dangerous spaces for humans to access.  Often, these tanks are filled with pipes, cables, beams and other structural elements that are critical to the operation of the vessel, but are usually not arranged in an optimal way to allow a human to easily move and navigate through the confined space.  Given these hazards, there is a lot of potential to use robots to perform many of these tasks. 

However, there are two major challenges in achieving a viable robotic solution for this problem. First, the interiors of these confined spaces are often discolored, poorly illuminated and unstructured, which cause issues for traditional vision-based localization, mapping and navigation approaches.  Second, the spaces are often irregularly shaped and cluttered with structural elements, as mentioned before.  Therefore, it is difficult even for a robot to move inside the tank and access all the components, rendering robust locomotion and precise manipulation completely non-trivial.  

A number of potential solutions have been explored to address the locomotion challenge, albeit for other types of confined spaces. For example, a quadruped climbing robot with a compliant magnetic foot has been developed to squeeze through entry portholes \cite{Bandyopadhyay2018}. Different robot designs have been investigated for in-pipe inspection, including a composition of active and passive compliant joints \cite{kakogawa2020}, snake locomotion patterns  \cite{Virgala2020} and adhesion-based crawling motions \cite{versatrax}. Various solutions have also been proposed to deal with the manipulation challenge. Representative examples include human-robot mixed-initiative control trading \cite{Owan2017} and task dynamics imitation learning \cite{Owan2020} for manufacturing inside confined spaces and a full stack autonomy framework for multi-task manipulation of irregular objects \cite{Han2020}.

We, instead, focus on the first challenge of robot localization, mapping and navigation inside confined spaces, and aim to decouple these capabilities from that of locomotion and manipulation.  In other words, our methods are designed to be used on any robot with on-board processing and payload carrying capacity, including those with novel locomotion and manipulation capabilities. In this connection, there are several related works, including semi-automated inspection of an industrial combustion chamber \cite{Tripicchio2018}; submerged building mapping by an autonomous vehicle \cite{Preston2018}; UAV-based localization in ballast tanks \cite{Brogaard2020}; intelligent exploration in mines using a small drone \cite{Akbari2020}; autonomous navigation through manhole-sized confined spaces using a collision-tolerant aerial robot \cite{DePetris2020}; and, semi-autonomous inspection of underground tunnels and caves \cite{Azpurua2021}.

We, however, address the specific task of detecting foreign object debris (FOD) left inside large water tanks. While the FOD detection problem has been studied widely in the literature, almost all the studies are for open spaces, especially airports, where the presence of such debris is particularly detrimental. Recent examples of airport FOD detection methods include multi-robot coordination \cite{Ozturk2016}, deep learning \cite{Cao2018,Gao2021}, and learning-based pixel visual features \cite{Jing2022} for standard optical cameras. Frequency modulated continuous mm-wave radars are often employed instead of cameras, where recent works include the use of fractional Fourier transform \cite{Lai2020}, power spectrum features-based classification \cite{Ni2020}, cross section characteristics \cite{Futatsumori2021}, line of sight visibility analysis \cite{Fizza2021}, adaptive leakage cancellation \cite{Liu2021}, and variational mode decomposition \cite{Zhong2021}. Alternate sensing approaches include the use of object minimal boundary extraction for infra-red cameras \cite{Kniaz2014}; and, scan or point cloud processing for light detection and ranging (LiDAR) sensors \cite{Mund2015,Elrayes2019}. 

Other open space FOD detection applications include inspection of aircraft damages \cite{Xu2018} and power transmission lines \cite{Zhang2019,Haotian2021}; quality control of graphics card assembly lines \cite{Kuo2022}; and real-time logistics monitoring \cite{Xiong2020} with visible light cameras. For confined space FOD detection, to the best of our knowledge, there has been only one reported work so far, where Latimer investigated processing of depth camera images for aircraft wing inspection \cite{Latimer2019}. On a somewhat related note, real-time detection of the differences of industrial parts from their corresponding computer-aided design (CAD) models has been done by processing the point clouds generated by hand-held laser scanners \cite{Kahn2013}.

Here, we present a new confined space FOD detection system using a local probabilistic outlier detection method on 
the visual maps generated by an autonomous ground robot equipped with just a standard depth camera. Note that we do not use any object detection or recognition method as the (expected) types or classes of objects are unknown in our case. Instead, we rely on remote human assistance, in the form of labeling of camera photos corresponding to candidate (likely) FOD regions, to achieve high detection accuracy. We also avoid using any expensive sensor such as 3D LiDAR, and show that our method works well with only visual odometry. These characteristics 
should make our system broadly applicable for various 
confined space inspection tasks
in large vehicles, such as aircraft, military tanks and spaceships, with hazardous operating conditions using robots employing different locomotion strategies. 

Specifically, the contributions of our work are three-fold:
\begin{itemize}[itemsep=0mm]
\vspace{-2mm}
    \item We develop a purely vision-based mapping system for human-assisted FOD detection in poorly-lit confined spaces.
    \item We devise a novel local Mahalanobis distance-based outlier detection method for point cloud representations of mapped spaces.
    \item We demonstrate promising FOD detection performance using both simulation and physical experiments on a ground robot exploring a representative confined space in the form of a water tank. 
\end{itemize}

\section{Preliminaries}
  
\begin{figure}[thpb]
      \centering
      \framebox{\parbox{0.8\textwidth}{ \includegraphics[width=0.8\textwidth]{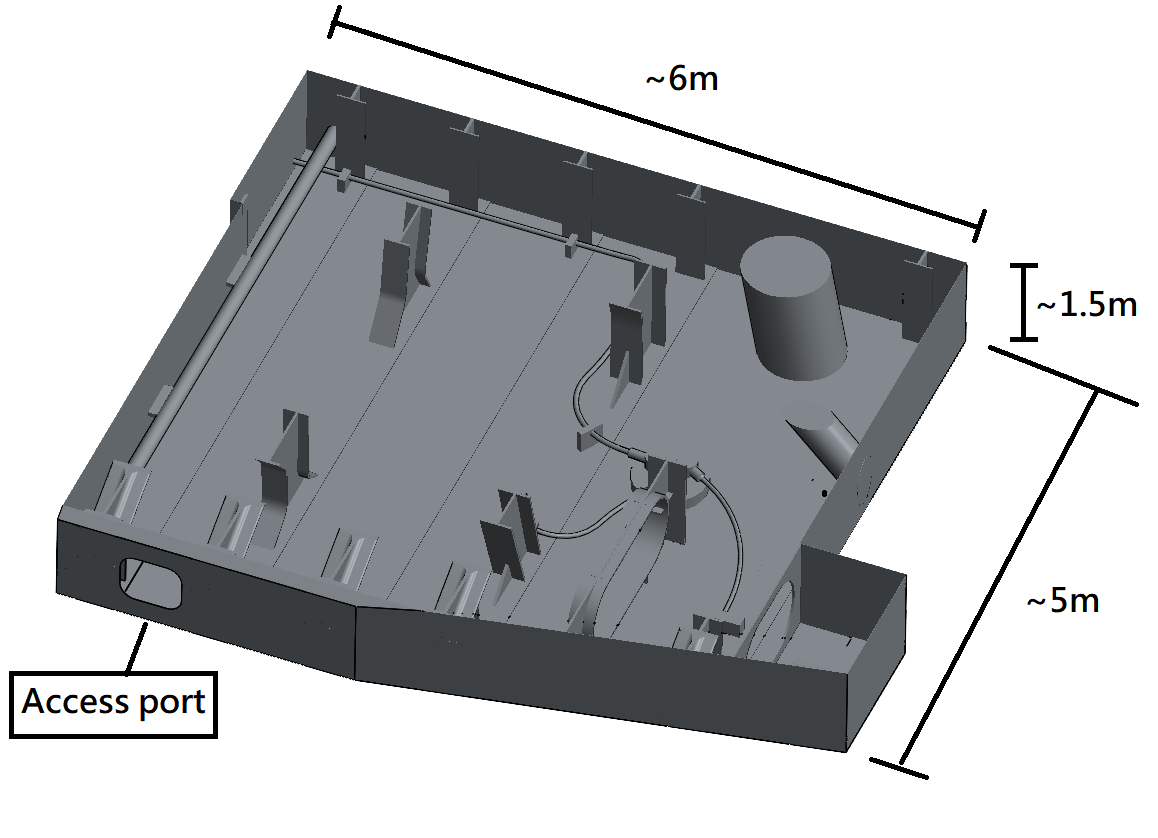}
}}
      \caption{CAD model of a generic water tank with the ceiling not shown to allow viewing of the interior components}
      \label{CAD}
   \end{figure}

A CAD model of a generic water tank was developed by the Naval Undersea Warfare Center (NUWC) Division Keyport to generate a realistic environment of a typical confined space found inside large vessels. The generic water tank model, as shown in Figure \ref{CAD}, included many beams, piping, cabling, and other structures commonly used in ship construction. Although this CAD model was intended to be an accurate representation of the challenging environment encountered during maintenance activities in confined spaces, it was not a precise replication of any existing vessel space. Therefore, this CAD model was publicly released and is widely available for use by other researchers\footnote{\url{https://github.com/blue-ring-octopus/FOD_Detection/blob/main/resource/GENERIC_WATER_TANK_DISTRO-A_NUWC_Keyport_Release_20-008.STL}}.

Technically, a FOD is any object that is alien to the environment, in this case the water tank. FOD has the potential to cause damages to the system during operation and must be removed via inspection. In a marine vessel, most common FODs include various standard hand tools, such as screwdriver and wrench, which are left behind after maintenance and repairs. In this paper, a FOD is simplified to include anything deviating from the given CAD model, which would primarily be a traditional FOD but could also include installation mismatch, tank defect or tank damage. 

\section{Technical Approach}
   
\begin{figure*}[thpb]
      \centering
      \framebox{\parbox{0.95\textwidth}{ \includegraphics[width=0.95\textwidth]{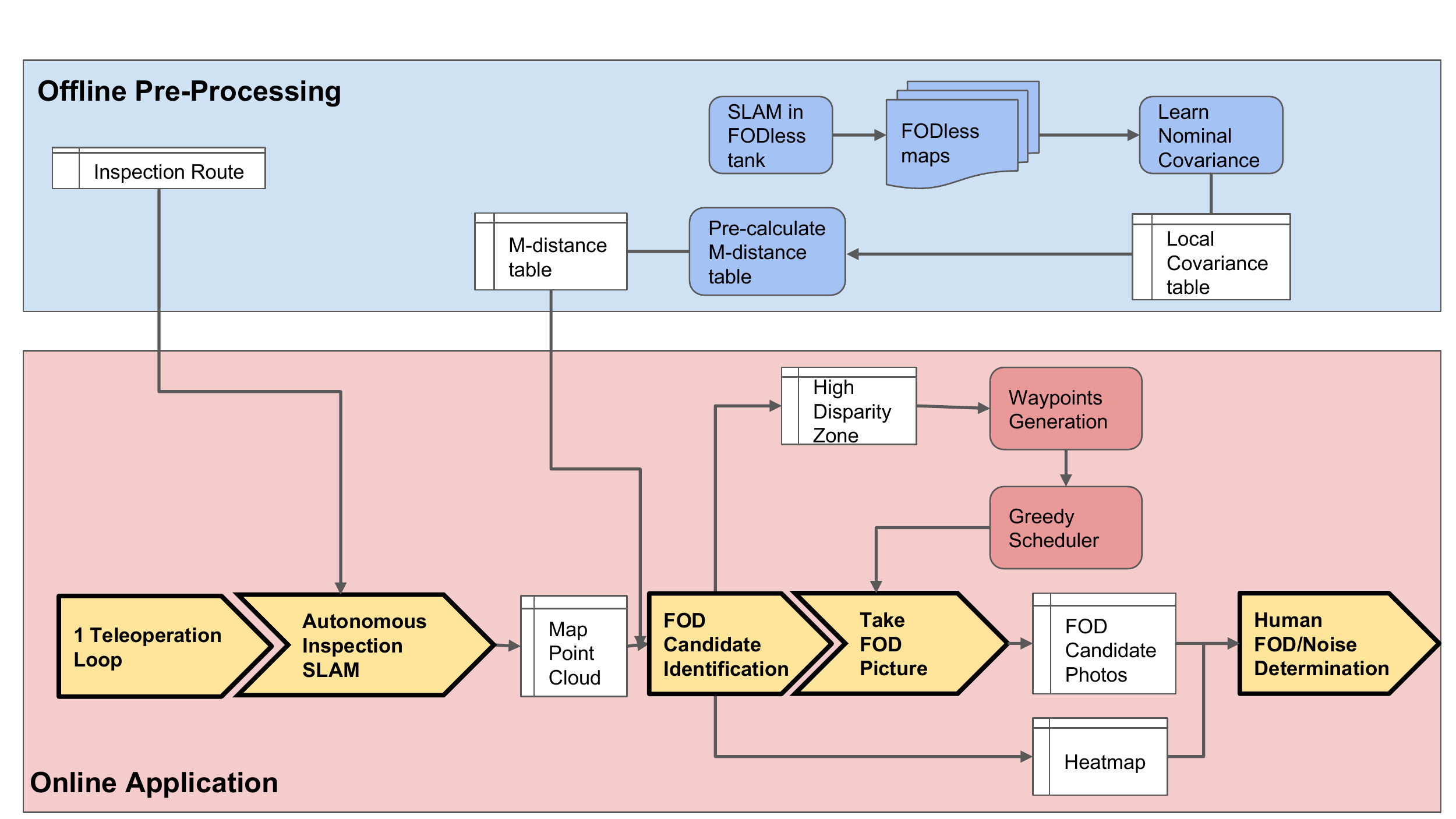}}}
      \caption{Overall pipeline of the FOD detection system}
      \label{pipeline}
   \end{figure*}

   To accomplish effective and efficient inspection, we incorporate human-in-the-loop 
   decision making to assist the semi-autonomous inspection system. The system is primarily running online under the Robot Operating System (ROS) framework, with various offline pre-processing modules to enhance the online performance. Pre-processing includes defining an inspection route for every inspection session; and generating a reference point cloud from either the CAD model or the collected nominal maps. The online inspection is divided into four major phases, as illustrated in Figure \ref{pipeline}. In the first phase, the robot collects the query point cloud by performing SLAM either via teleoperation or autonomous navigation using the predefined inspection route. The second phase begins when the SLAM is completed. The point cloud from the visual SLAM is exported and compared with the reference point cloud. Points with high discrepancy are segmented and clustered into FOD candidates. In the third phase, the centroid of each candidate is projected to the navigation cost map to determine the waypoints from where the robot expects to see the candidate cluster. The robot then covers all the waypoints, selecting the one closest to its current location as the next destination, and takes (camera) pictures of the FOD candidates. In the final phase, the pictures taken by the robot are presented to a human to determine whether FODs are present.
  
\subsection{SLAM and Navigation}

The SLAM functionality is provided by RTAB-Map via ROS \cite{rtabmap}. The main configuration follows the default launch file from rtabmap\_ros\footnote{\url{https://github.com/introlab/rtabmap_ros/blob/master/launch/rtabmap.launch}}, with the modified parameters and nodelets shown in \ref{rtabmap_param}. The main objectives for the SLAM are self-localization in the confined space, and construction of a point cloud for each inspection session using a depth camera. RTAB-Map is chosen based on its open source ROS implementation, a wide variety of sensor compatibility, and good reported performance on many different tasks. In addition, RTAB-Map has a modular approach toward odometry, which allows a simple switch between wheel odometry, in-built visual odometry, or any other third party odometry algorithm without affecting the SLAM functionality. 

The navigation functionality is provided by the ROS navigation stack in the TurtleBot3 package. The navigation stack takes the 2D occupancy grid map generated by the SLAM package as an input and creates a 2D occupancy grid cost-map in real-time. Once a waypoint is published to the waypoint topic, the optimal trajectory from the current location to the specified waypoint is calculated and executed through velocity control. 

\subsection{FOD Candidate Identification}
\begin{figure}[thpb]
      \centering
     \framebox{\parbox{0.75\textwidth}{ \includegraphics[width=0.75\textwidth]{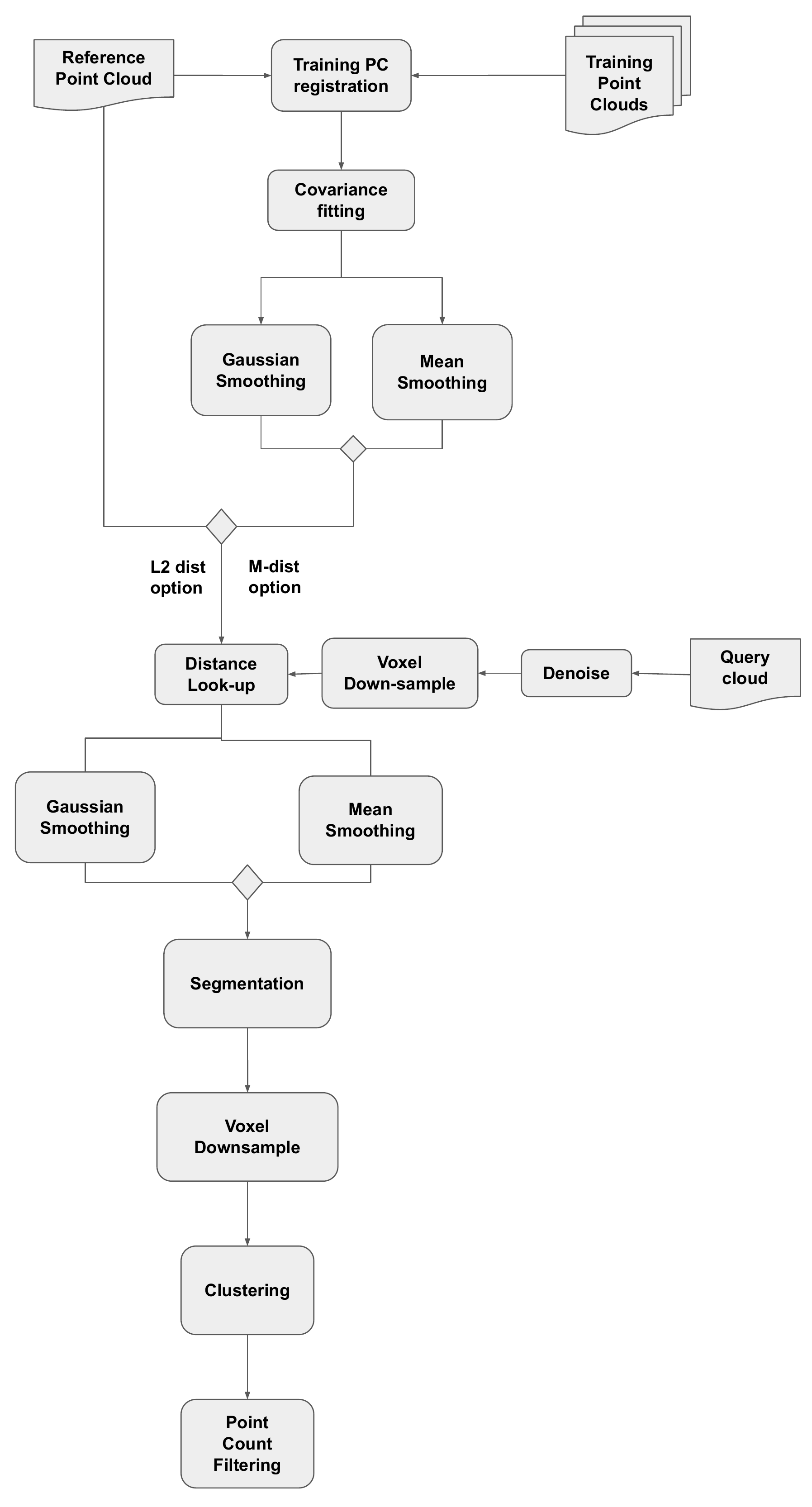}}}
            \caption{FOD candidate identification procedure}
      \label{fod_id}
   \end{figure}
   
   \begin{figure}[thpb]
      \centering
     \framebox{\parbox{0.9\textwidth}{ \includegraphics[width=0.9\textwidth]{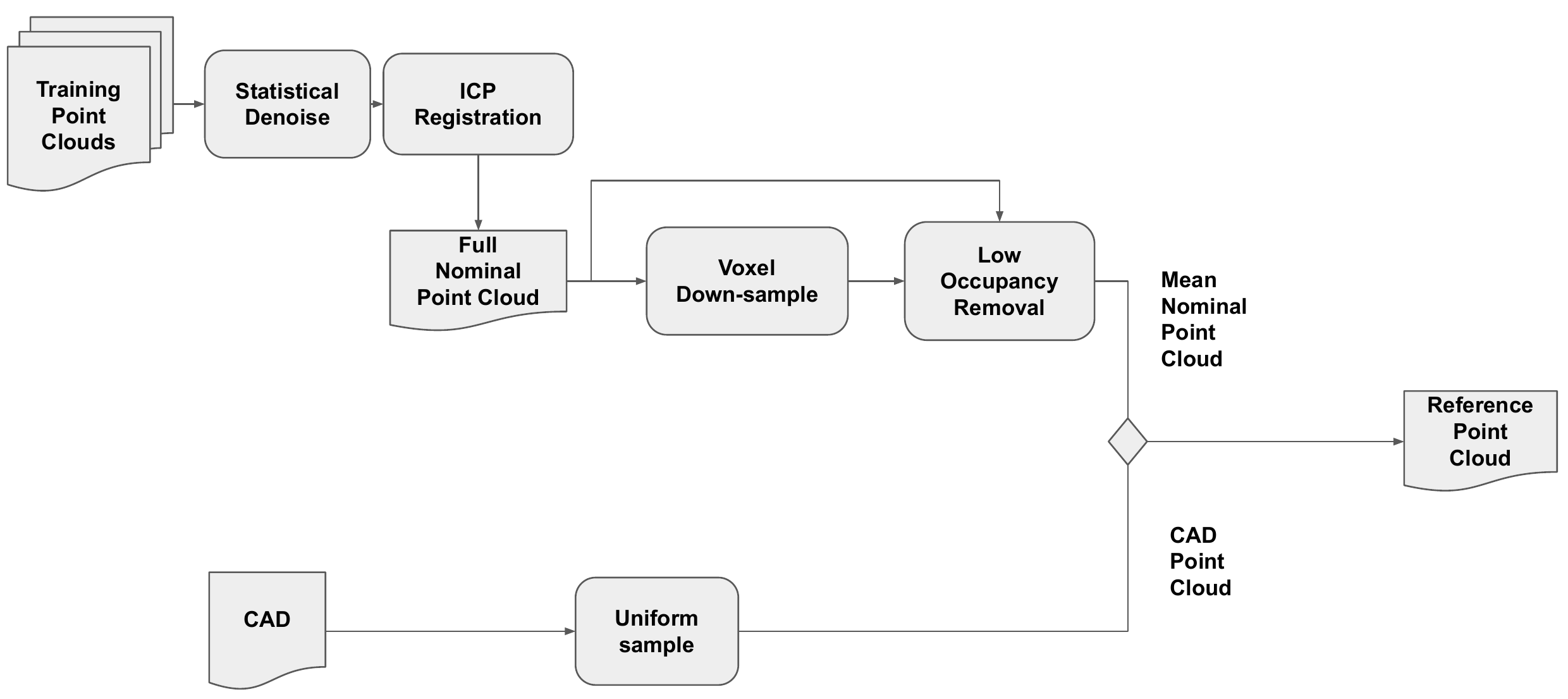}}}
            \caption{Reference point cloud generation procedure}
      \label{ref_gen}
   \end{figure}

We develop our own FOD candidate identification algorithm to detect potential foreign objects from the SLAM point cloud while preventing misclassification due to noise. The core idea is to compare the \textit{discrepancy} of each point in the point cloud to a nominal reference map of the environment. The overall pipeline for the identification process with the various options is shown in Figure \ref{fod_id}. 

 \subsubsection{Reference Point Cloud Generation}
  The reference point cloud is a point cloud representation of the water tank that is known to be FOD-free. It is treated as the ground truth and any points that deviate from it by more than the cutoff value are considered FOD candidate points. Here, we consider two options for reference point cloud generation. It is created either using a mesh of the tank CAD model or by collecting sample point clouds in the FOD-free tank. Figure \ref{ref_gen} shows the procedure of generating the reference point cloud using both the options.

        The CAD mesh method assumes that a accurate model of the environment is available, wherein, the CAD model is first exported as a PLY format triangle mesh and the mesh is uniformly sampled to create a dense point cloud. The sample point clouds method, on the other hand, is developed for use in a confined space with a large number of unmodeled structures and/or construction variations from its CAD model. In this case, all the sample point clouds from the inspection training sessions are first de-noised and then registered, either with the CAD model or one of the sample clouds. Next, all the sample point clouds are merged into a single point cloud. The merged point cloud is then voxel down-sampled to calculate the mean points occupying a single voxel. This down-sampling process merges all the points in a voxel into a single point regardless of the number of points. However, such a merger sometimes results in noisy points creating false voxel occupancy conditions. To eliminate this effect, all the points generated from a voxel with occupancy counts less than a threshold quantile are removed.
        
\subsubsection{Covariance Fitting}
    The discrepancy metric is a scalar value assigned to each point in a point cloud. The metric scales according to the estimated amount of deviation of these points to their corresponding points in the reference point cloud. Euclidean distance for nearest neighbors is used as the \textit{de facto} metric while comparing the two point clouds. In our case, we consider the nearest neighbors of the points in the query point cloud to the nominal point cloud. However, this consideration often leads to noisy mapping of the walls and beams with high deviation values and either a) cause a large number of FOD candidates for a low distance threshold; or, b) are insensitive to small sized FODs for a high distance threshold.
         
         We address this issue by using a probabilistic approach based on local \textit{Mahalanobis distance} (M-distance). Alternatively, a spatial Chi-squared test can be used for local outlier detection \cite{cloudcompare}; however, it works well only for homogeneous density point clouds. 
        \begin{equation}
            D_M=\sqrt{(x-\mu)^T\Sigma^{-1}(x-\mu)}  .
            \label{M-dist}
        \end{equation}
        The M-distance $D_M$, shown in (\ref{M-dist}), introduces an expected noise in the form of the covariance matrix \(\Sigma\). It also adds directionality to the expected noise, which is especially useful in distinguishing the walls with horizontal deviations and beams with vertical deviations. While \(\mu\) is defined as the closest point from the training or query cloud point $x$ to the reference point cloud, \(\Sigma\) is estimated from the training samples collected for the mean point cloud. If the CAD model is used for reference point cloud generation, a set of training point clouds has to be still collected for \(\Sigma\) estimation.
        
        Instead of estimating one single global covariance matrix for the whole tank, we compute local covariances. It is done by first calculating the signed spatial error
        \begin{equation}
             \Delta x_{i,j}=\begin{pmatrix}
                                x_j\\ 
                                y_j\\ 
                                z_j
                                \end{pmatrix}-\begin{pmatrix}
                                x_i\\ 
                                y_i\\ 
                                z_i
                                \end{pmatrix}
        \end{equation}
        from each point \(x_j\) on the sample point clouds to their nearest neighbor point \(x_i\) on the reference point cloud. Next, for each point on the reference point cloud that contains at least one sample point, we calculate the scatter matrix \(S_i\)  as:
        \begin{equation}
            S_i=\sum_j(\Delta x_{i,j})\cdot(\Delta x_{i,j})^T .
            \label{scatter}
        \end{equation}
        If this quantity is divided by the number of samples, 
        we get the maximum likelihood estimation (MLE) of the covariance for that point on the reference map \(\Sigma_i\). It is, however, deliberately left undivided to smooth out the covariance matrices within a local region. Otherwise, 
        depending on the number of training maps, the point density of each training map and the voxel size of the nominal map, we risk having a large number of points on the nominal map with zero sample size and many points with a low sample size. 
        
        We consider two options for smoothing: mean smoothing
        and Gaussian smoothing.
        For mean smoothing, the covariances, $\Sigma_{i,m}$, are calculated regardless of the distance of the neighboring points to the center point
        using the formula:
        \begin{equation}
           \Sigma_{i,m}=\frac{1}{\sum_{j=1}^k n_j}\sum_{j=1}^k S_j .
        \end{equation}  
        Here, $k$ is the number of neighbors around the $i$-th point on the nominal map, \(n_j\) is the numbers of samples of the neighbors, and \(S_j\) is the scatter matrix of the $j$-th neighbor. For Gaussian smoothing, the covariances, $\Sigma_{i,G}$, are computed similarly but with the distances of the neighbors modeled as reliability weights \cite{reliability_weight}. This weighting method provides a non-frequency based unbiased estimator of the sample covariance.  
        The weights are based on a Gaussian kernel and a tunable roll-off rate parameter $\sigma$ as:
        \begin{equation}
            w_j=\exp{\left [  {-\frac{\left \|\Delta x_{i,j}  \right \|^2}{\sigma^2}}\right ]}
        \end{equation}
        \begin{equation} V_1=\sum_{j=1}^{k}n_j w_j \quad
            V_2=\sum_{j=1}^{k}n_j w_j^2
        \end{equation}
        \begin{equation}
          \Sigma_{i,G}=\frac{1}{V_1-V_2/V_1}\sum_{j=1}^kw_jS_j.
        \end{equation}
        To define the smoothing neighborhood, we consider two alternatives: $k$-nearest neighbor and spherical region of interest. 
        
            \begin{figure}[thpb]
              \centering
              \framebox{\parbox{0.8\textwidth}{ 
             \includegraphics[width=0.8\textwidth]{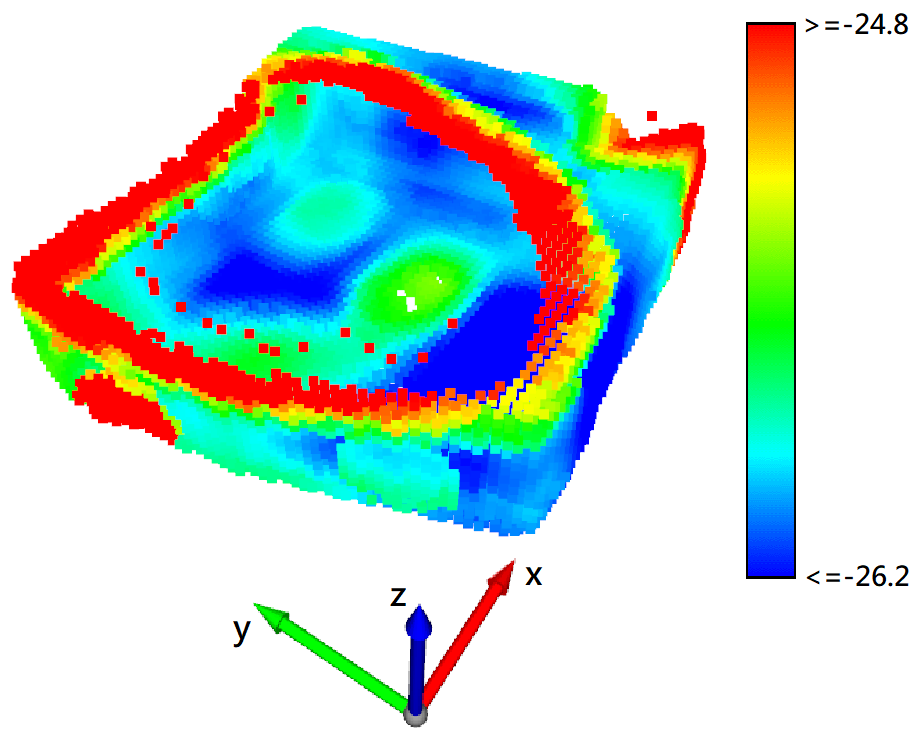}}}
              \caption{Heat map of the log-determinant of the local covariance matrix in the mean point cloud of the FOD-free water tank, with the minimum value clamped to $10^{th}$ percentile and the maximum value clamped to $90^{th}$ percentile}
              \label{covmap}
            \end{figure}
        
        Prior to smoothing, an optional voxel down-sampling step can be performed on the reference point cloud to reduce the computational burden. In that case, a nearest neighbor search is performed to find the closest point from the full point set to the down-sampled point set. The covariances for the points in the full point set are then simply set to the covariances of their nearest neighbors. Figure \ref{covmap} shows an example heat map of the log-determinant of the local covariances in the tank using sample point cloud and mean smoothing. As expected, the highest covariance values occur at the ceiling and behind the column, both of which are not adequately observed by the robot. The access hole area also has high covariance as the tank cover is manually placed, leading to some variations in its location among the different trials.
        
    \subsubsection{Discrepancy Query}
    Once the reference point cloud is generated and the covariances are estimated, a discrepancy metric is applied to query point clouds to identify the (candidate) FODs. Instead of directly computing the metric on the raw point cloud, several steps are applied to reduce noise and improve computation speed. First, a basic statistical outlier removal method is applied to to remove the noisy points that are far away from its neighbors in the query point cloud. A voxel down-sampling is then applied to the point cloud to keep the point count at a manageable level. The number of points corresponding to each voxel, \(n_i\), is also saved as a weight for future use to avoid losing density information from the down-sampling process. 
    
    The query process is similar to the training process. The associated point \(x_j\) on the reference point cloud is drawn for each voxel point \(x_i\) in the query cloud based on the shortest Euclidean distance as the discrepancy metric for the baseline L2-distance method. For the M-distance method, the discrepancy is found with a local version of (\ref{M-dist}) as
    \begin{equation}
        d_M=\sqrt{(x_i-x_j)^T\Sigma_j^{-1}(x_i-x_j)} ,
        \label{M-dist_local}
    \end{equation}
    where \(\Sigma_j\) is the covariance matrix of the point \(x_j\). Effectively, the noisier (higher covariance) a region is, the smaller is its discrepancy for the query point. For example, in Figure \ref{covmap}, the discrepancy of a point with a large distance value near the ceiling is scaled down by the high covariance of the ceiling; a point with the same distance but near the ground has a greater discrepancy due to the smaller covariance of the ground. To reduce high discrepancy values from sparse noisy points, a scalar version of the smoothing used during covariance fitting is applied on the discrepancy metric for each point. Mathematically, it amounts to substituting \(n_i d\) for \(S_i\) in (\ref{scatter}), where $d$ is the discrepancy metric. 

\begin{figure}[thpb]
      \centering
      \framebox{\parbox{1.0\textwidth}{ \includegraphics[width=1.0\textwidth]{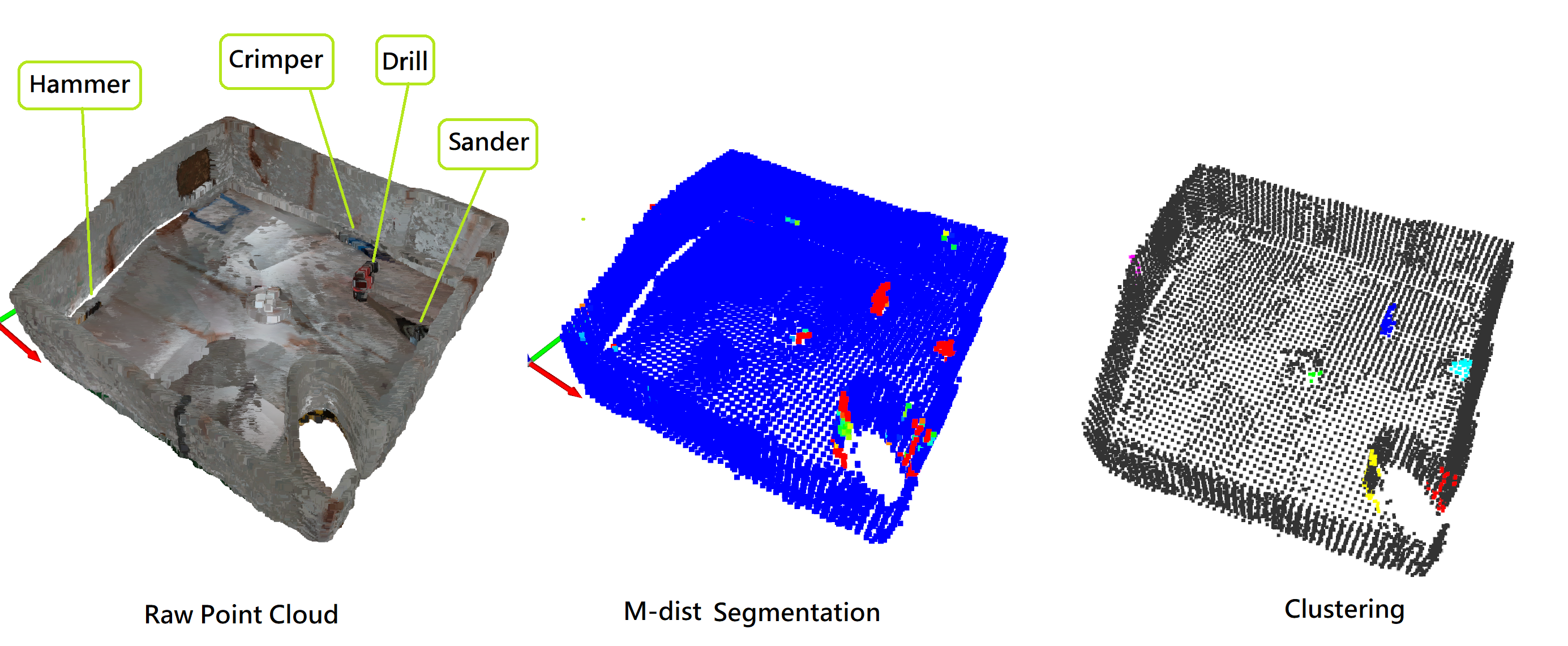}}}
      \caption{An example query point cloud with the high M-distance points segmented and clustered into different candidate FOD clusters with each cluster shown using a distinct hue}
      \label{fod}
   \end{figure}

        Every point with a discrepancy higher than the threshold is segmented out from the query point cloud and clustered into FOD candidates using hierarchical clustering. The points are agglomerated based on the Euclidean distance to the centroid of the existing clusters. Two clusters are merged if the distance between them is lower than a cutoff value, and any cluster with a fewer number of points than a minimum acceptable point count is rejected from the FOD candidate set. The selection of the threshold, clustering cutoff, and minimum acceptable point count are discussed in the Experiments section. The centroid of each cluster is then sent to the waypoint generation algorithm to proceed to the photo taking phase. Figure \ref{fod} shows an example of segmented and clustered FOD candidates point cloud, with the points lower than the threshold colored dark grey and each FOD candidate cluster colored using a different hue.
        
\subsection{Waypoint Generation}
    Once the FOD candidate locations are computed, they are passed to the waypoint generation module. First, a two dimensional occupancy grid map is created 
    for the query point cloud. 
    To prevent waypoints generation in unreachable locations, a flood fill is performed using the robot location as the seed value. The flood filled image is then subtracted by the original image to remove the walls, and inverted to obtain zero occupancy in the obstacle-free interior of the water tank. The occupancy grid is then inflated into a cost-map according to the ROS navigation stack \cite{costmap}.
    
\begin{figure}[thpb]
  \centering
  \framebox{\parbox{0.9\textwidth}{ \includegraphics[width=0.9\textwidth]{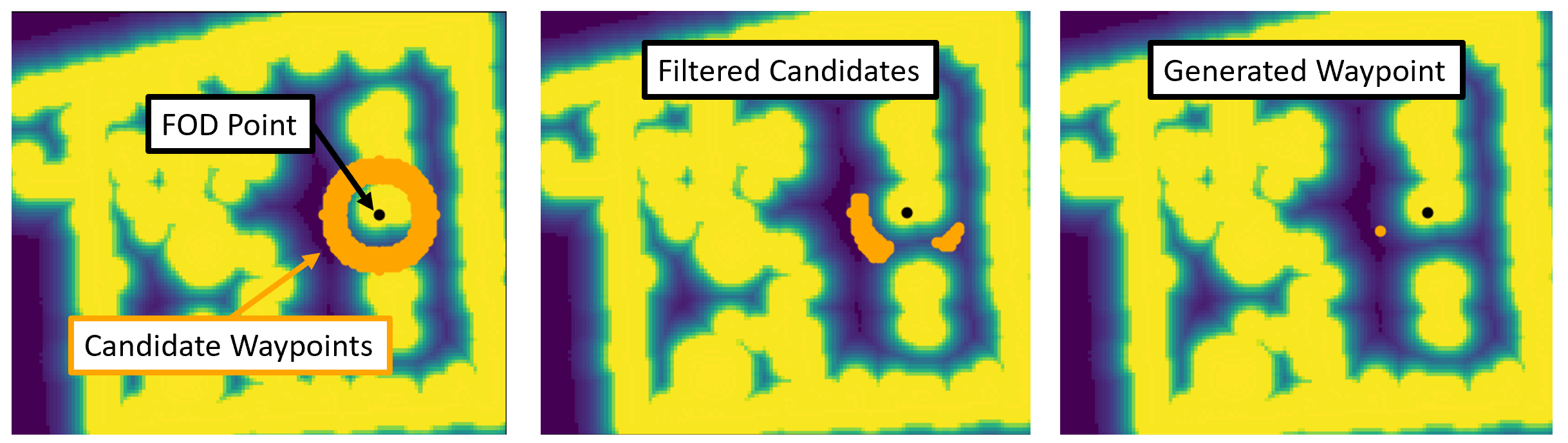}}}
  \caption{Waypoint generation procedure given FOD candidate location: (a) 2D occupancy grid with yellow pixels representing obstacles and purple pixels denoting free space (b) Flood fill to remove unreachable locations outside the tank (c) Cost map generated from the occupancy map (d) Sampling of waypoint candidates along a ring centered at the candidate FOD cluster (e) Filtering of waypoint candidates based on the cost map (f) Final waypoint selected using ray tracing}
  \label{waypoint_gen}
\end{figure}

    Candidate waypoints are created surrounding the FOD location within a minimum and maximum range so as to provide an acceptable image of the FOD. The candidate waypoints are then filtered to avoid colliding with the surroundings and ensure that the FOD is visible. To avoid collisions, the waypoints that overlap with high-cost regions of the cost-map are removed. To ensure that the FOD is visible, each remaining waypoint casts a ray between itself and the FOD. The ray is terminated if it collides with the surroundings in the FOD-less cost-map, and the corresponding waypoint is removed. Each ray records its cumulative cost, which is the sum of the cost in the grid cell under the ray at each time step. The candidate waypoint with the least cost is chosen as the final waypoint. Figure \ref{waypoint_gen} shows an example of the waypoint generation process. 
    

\section{Experiments}
The experiments were conducted in two stages.  Preliminary testing was done in simulation, followed by extensive physical (hardware) trials on a scaled-down tank prototype. Human subjects studies are included in both the experiment stages. 

\subsection{Simulation}
The simulation was done using Gazebo in Ubuntu with ROS Melodic as the framework and Python as the primary programming language. The CAD model of the tank was painted with a rusty white texture 
and exported as a DAE file using Blender. The DAE file was spawned in Gazebo with all the natural light sources disabled to recreate the dark confined space environment. TurtleBot3 Waffle Pi was chosen as the robot model. The original Pi camera was replaced by two identical cameras using Intel RealSense D435's stereo camera specifications without the active infrared projector. A spot light source was added to the front of the robot to act as a flashlight.

The experiments comprised a total of 30 trials, with 15 trials using the RTAB-Map's built-in visual odometry and another 15 trials using the robot's wheel odometry to study the viability of pure visual odometry. For each trial, the robot was spawned near the access hole with the same pose. Two to five FODs were randomly spawned with random poses from a pool of six FOD types, consisting of drill, screw driver, hammer, wrench, level, and sander. Point cloud registration was performed using the Open3D iterative closest point (ICP) method. All the parameters were manually chosen. Random down-sampling with a ratio of 0.1 was used in place of voxel down-sample. A CAD model was used as reference point cloud and no denoising was performed. M-distance was used as the metric with Gaussian smoothing ($\sigma=0.05$), a spherical region of interest (radius of $4\sigma$), a threshold distance of 1.75, and clustering cutoff of 0.275.

The resultant FOD photos were assembled into online questionnaire surveys, with each survey containing all the photos from a single trial. The administration of the anonymous survey was approved by the University of Washington (UW) Institutional Review Board with the study \# STUDY00013902. The surveys were sent out to UW students and Naval Undersea Warfare Center (NUWC) Division Keyport personnel. A total of 23 responses were received, of which 61.9\% were engineers, 33.3\% graduate students, and 4.8\% managers. 57.1\% were in the 18-30 age group, 28.6\% in the 31-45 group, and 14.3\% in the 46-60 group. 81.0\% of the participants were males and 19.0\% were females. 

\begin{figure}[thpb]
      \centering
      \includegraphics[width=0.85\textwidth]{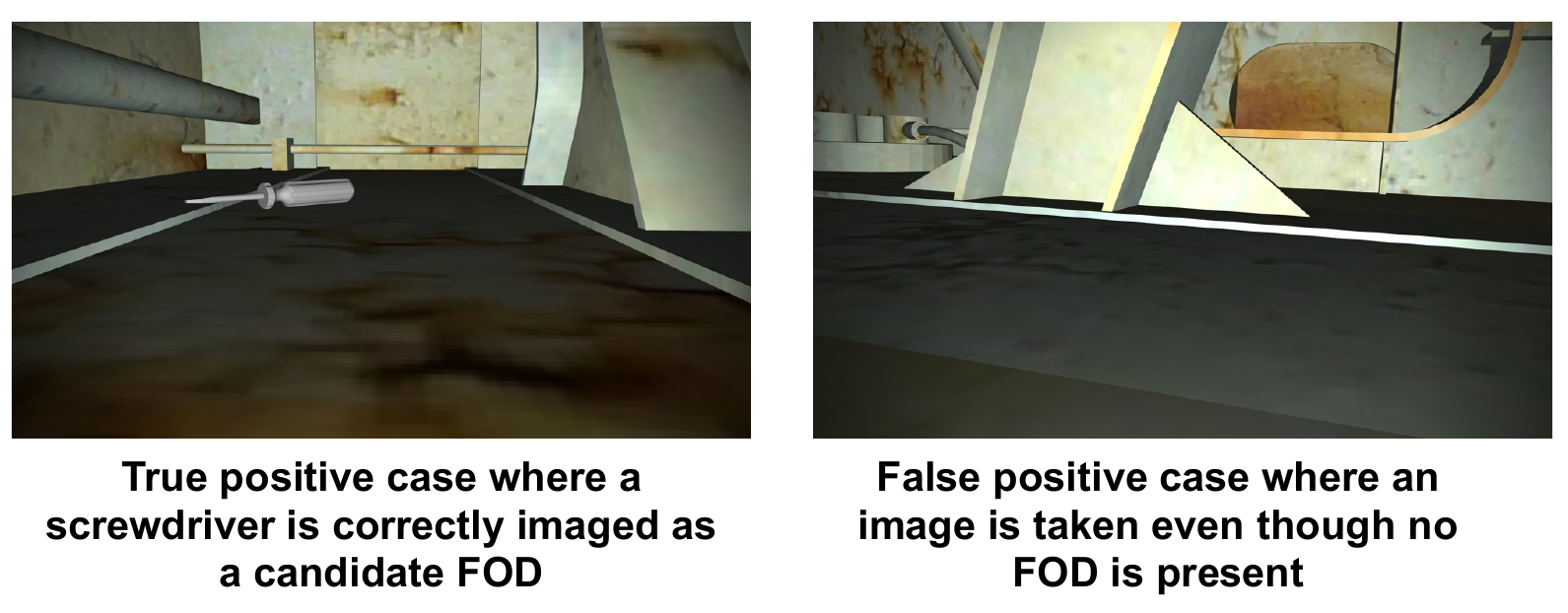}
            \caption{Examples of candidate FOD photos, taken by the inspection robot, that are shown to remote humans for final decision-making.}
      \label{FOD_candidates}
   \end{figure}

We first analyzed the performance of the FOD detection approach before looking at the effectiveness of remote humans in making the final decisions. The relevance of the FOD photos is shown in Table \ref{Relevance}, where the ``photo contains FOD'' category includes the same FOD appearing in multiple photos and photos containing partial FOD images. The precision is defined as the number of photos containing FODs divided by the total number of photos. It is less than 50\% for both the odometry methods, with wheel odometry being slightly lower than visual odometry. These low values are a direct result of high sensitivity by choosing a low M-distance threshold. Figure \ref{FOD_candidates} shows an example of a photo containing FOD, with a gray screwdriver at the center of the image, and 
an example of a photo with no FOD, which is a false positive detection due to the noise associated with the I-beam structure. 

\begin{table}[h]
\caption{Relevance of Candidate FOD Photos in a Simulation Study}
\vspace{2mm}
\begin{tabular}{|c|c|c|c|}
\hline
\textbf{}                &  \textbf{Contains FOD} & \multicolumn{1}{l|}{\textbf{No FOD}} & \multicolumn{1}{l|}{\textbf{Photo Precision}} \\ 
\hline
\textbf{Wheel Odometry}  & 64                     & 104                                          & 0.381                                            \\
\textbf{Visual Odometry} & 77                     & 109                                          & 0.414                                            \\
\hline
\end{tabular}
      \label{Relevance}
\end{table}

   

The effect of high sensitivity, or recall, is shown in Table \ref{rate}. The first column shows the total number of FODs detected by the photo set, which is the number of FODs present in a session and captured in at least one photo. This differs from the photo containing FOD in that only distinct FODs are counted here.
The second column shows the number of missing FODs, which are FODs present in a session but not seen in any of the photos. 
The detection rates are high as previously discussed, with both the odometry methods having recall greater than 90\%. This suggests that if the human are able to spot all the FODs in the photos, over 90\% of the FODs can be detected and removed afterward. 

   \begin{table}[h]
\centering
  \caption{FOD Detection Rate in a Simulation Study}
  \vspace{2mm}
\begin{tabular}{|c|c|c|c|}
\hline
\textbf{}                &  \textbf{Detected FOD} & \multicolumn{1}{l|}{\textbf{Missed FOD}} & \multicolumn{1}{l|}{\textbf{FOD Recall}} \\ \hline
\textbf{Wheel Odometry}  & 51                     & 3                                         & 0.944                                           \\
\textbf{Visual Odometry} & 48                     & 4                                         & 0.923                                            \\
\hline
\end{tabular}
      \label{rate}
\end{table}

 The accuracy of identifying the exact FOD type comes out to be 0.860. The confusion matrix for exact FOD labeling is reported in Table \ref{con_mat_id}. Unlike standard confusion matrices, the labels contain a ``not sure'' option for participants that find a FOD but cannot identify it. We also have a ``mixed FOD'' column to include photos with multiple detected FODs and multiple mislabeled FODs, since it is not possible to deduce which FOD is causing the confusion. As before, the matrix shows a high labeling accuracy with the maximum values in the diagonal entries. 

\begin{table}[h]
\centering
  \caption{Confusion Matrix for Human Labeled FOD Photos over 30 Simulation Trials}  
  \vspace{2mm}
\resizebox{\columnwidth}{!}{
\begin{tabular}{cc|cccccccc|}
\cline{3-10}
                                                    &                      & \multicolumn{8}{c|}{Actual Type}                                                                                                                                       \\ \cline{3-10} 
                                                    &                      & \multicolumn{1}{c}{\textbf{Hammer}} & \textbf{Level} & \textbf{Screwdriver} & \textbf{Wrench} & \textbf{Sander} & \textbf{Drill} & \textbf{No FOD} & \textbf{Mixed FOD} \\ \hline
\multicolumn{1}{|c|}{\multirow{8}{*}{\rotatebox[origin=c]{90}{Labeled Type}}} & \textbf{Hammer}      & 6                                    & 0              & 0                    & 0               & 0               & 0              & 0               & 1                  \\
\multicolumn{1}{|c|}{}                              & \textbf{Level}       & 0                                    & 17             & 0                    & 0               & 0               & 0              & 0               & 0                  \\
\multicolumn{1}{|c|}{}                              & \textbf{Screwdriver} & 0                                    & 0              & 33                   & 0               & 0               & 0              & 0               & 0                  \\
\multicolumn{1}{|c|}{}                              & \textbf{Wrench}      & 0                                    & 0              & 0                    & 20              & 0               & 0              & 0               & 0                  \\\multicolumn{1}{|c|}{}                              & \textbf{Sander}      & 0                                    & 0              & 0                    & 0               & 2               & 0              & 0               & 0                  \\
\multicolumn{1}{|c|}{}                              & \textbf{Drill}       & 0                                    & 0              & 0                    & 0               & 0               & 11             & 0               & 0                  \\
\multicolumn{1}{|c|}{}                              & \textbf{No FOD}      & 0                                    & 0              & 1                    & 4               & 6               & 0              & 163             & 0                  \\
\multicolumn{1}{|c|}{}                              & \textbf{Not Sure}    & 2                                    & 13             & 0                    & 8               & 1               & 2              & 3               & 0                  \\ \hline

\end{tabular}
}
      \label{con_mat_id}
\end{table}  
 
\begin{figure}[thpb]
      \centering \includegraphics[width=0.85\textwidth]{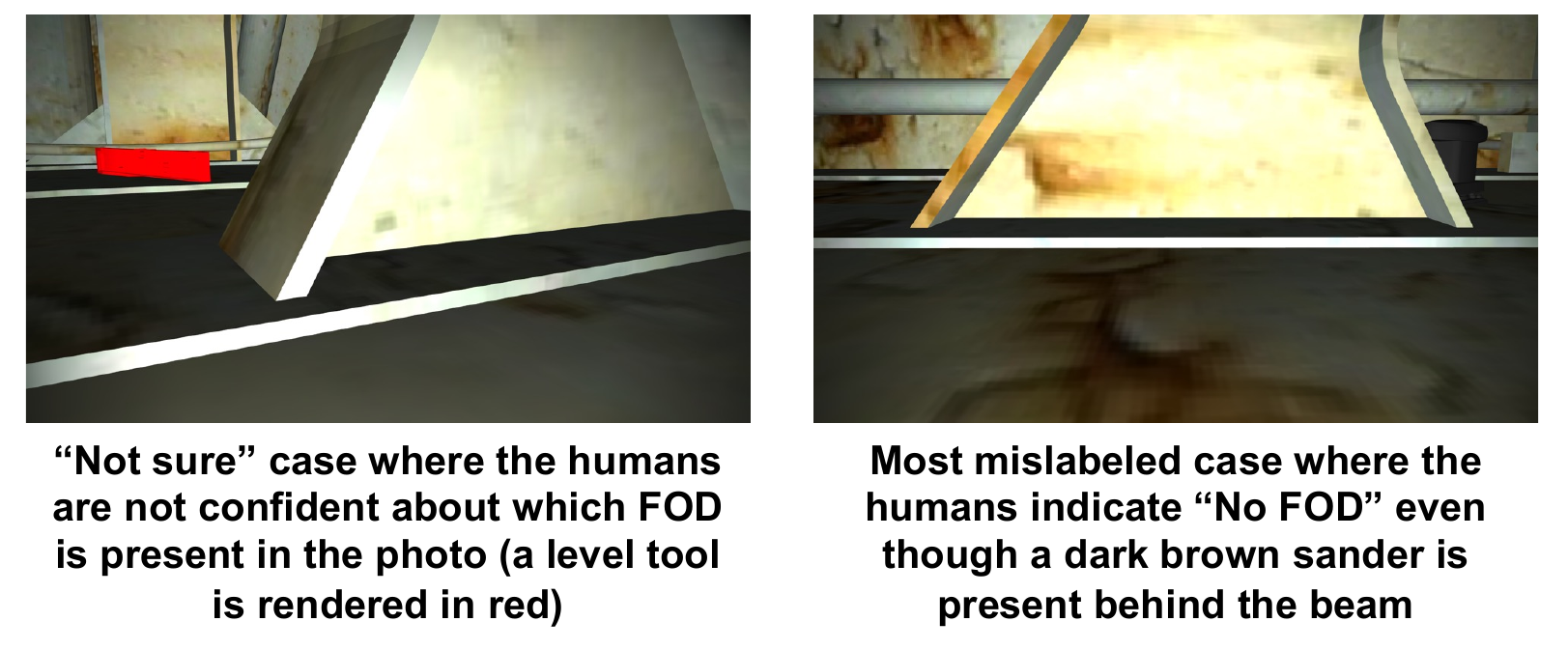}
      \caption{Candidate FOD photos causing maximum confusion to the human subjects during online questionnaire survey}
      \label{Confusion}
   \end{figure}

 As reported in Table \ref{con_mat_id}, the spirit level is the most confused FOD, which is likely due to a low-quality rendering of the level. Figure \ref{Confusion} shows an example of such a confusing photo, with the level colored uniformly red. While the color red is easy to spot in the photos, the detailed shape is hard to see with uniform coloring, which makes the level appear as a red bar from a distance. On the other hand, the item most commonly mislabeled as ``No fod'' is the sander. Similar to the spirit level, the sander has a less distinct outline as compared to the other FODs. However, unlike the bright red color of the level, the sander is colored dark gray, as shown in Figure \ref{Confusion}, which makes it similar to the hue of the rusty tank texture. Hence, it is quite easy for the participants to completely miss the sander. 
 
 These rendering problems are, however, unique to simulation and should not be a concern for real-world deployment, as evident from the physical trials results reported next. More importantly, we observe promising results on detecting FODs using just visual odometry, where both precision and recall are almost identical to that using accurate wheel odometry. Consequently, we choose visual odometry during the physical trials to demonstrate the potential of our detection approach even for non-wheeled robots.

\subsection{Physical Trials}
\subsubsection{Setup}
\begin{figure}[thpb]
      \centering \framebox{\parbox{0.7\textwidth}{\includegraphics[width=0.7\textwidth]{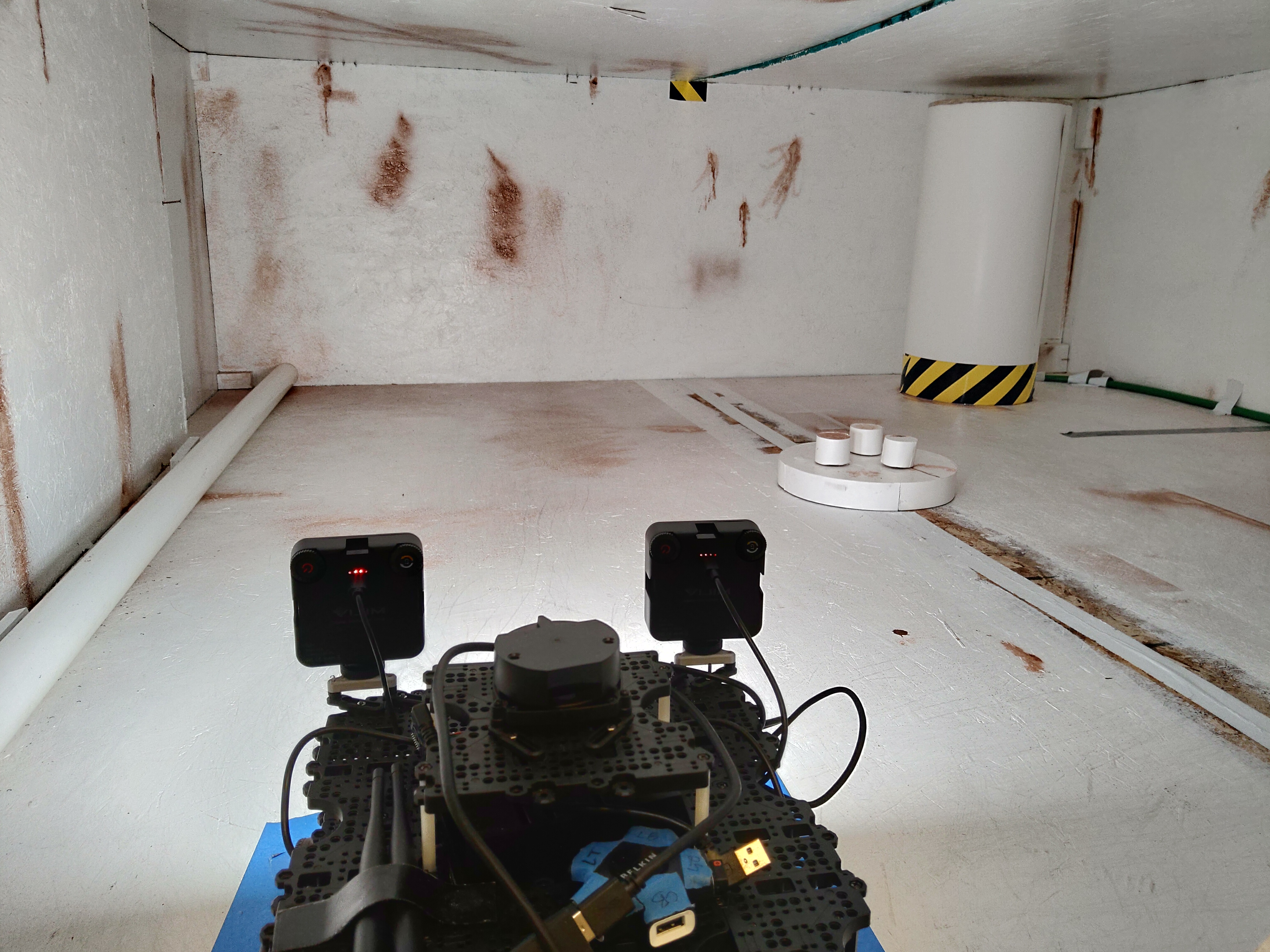}}}
      \caption{Nominal Physical Tank Configuration}
      \label{tank}
  \end{figure}

A modified TurtleBot 3 Waffle Pi was used as the robot platform. Instead of using the baseline Raspberry Pi 3B+, an Nvidia Jetson AGX Xavier was added to the robot to provide on-board GPU capability. The default Pi camera was also replaced by a RealSense D435i depth camera. Two portable photography lights were mounted on top of the robot to reduce strong specular reflection from the ground. A scaled down prototype of the water tank was built from ply wood and painted with a white base with rusty brown spots to mimic the actual tank texture. Figure \ref{tank} shows an image of the FOD-free tank taken from the access hole with the robot at the starting location. 

   All the computation was done in Python 3.7, with several methods parallelized using the CUDA library from Numba 0.53.1. The mean point cloud option was chosen for reference point cloud generation as the physical tank had significant variations from the CAD model. The voxel size of the mean point cloud and the occupancy threshold were chosen to be 0.05 m and $25^{th}$ percentile, respectively. For down-sampling, we used ``voxel down-sample and trace'' to handle stochasticity with a voxel size of 0.02 m. 
   For smoothing, we employed the $k$-nearest neighbors (mean) method to avoid numerical instabilities from the estimated Gaussian weights for large distances and poor performance from searching spherical neighborhoods for highly variable density point clouds. $k$ was selected as 250 for covariance smoothing and 50 for discrepancy smoothing. These parameters were manually tuned using the validation set discussed below. A grid search was used to obtain the threshold and cutoff parameters, which depended on the choice of the discrepancy metric, as discussed in the next section. 

15 nominal training maps were collected to generate the mean point cloud and fit the local covariances. In addition, 15 validation maps with FODs were collected for parameters tuning. The FODs were randomly sampled from the following set: hammer, power drill, tape measure, screw driver, sander, and crimper. The FOD samples and their locations were generated as follows. First, the total number of FODs was randomly chosen between 2 to 5. The corresponding number of FODs were then uniformly sampled without replacement from the full FOD set. For each chosen FOD, the $x,y$ location was uniformly sampled from the bounding box of the mean point cloud and $z$ was fixed to the ground plane; if the location was sampled outside the tank, the FOD was placed at the closest interior location. Another 15 FOD-containing test maps were generated in a similar manner. The FOD centroids were adjusted according to their actual locations in the registered point clouds for use as ground truth values.

\subsubsection{Parameter Search}

The optimal values for the discrepancy threshold, clustering cutoff, and minimum point count were obtained using a grid search method over a set of 15 validation (mapping) trials. 30 cluster cutoff values were generated with a linear spacing from 0.05 to 1. 20 discrepancy values were generated, with the lower bound of the linear space as the minimum $25^{th}$ percentile of the discrepancies over all the 15 trials. The upper bound was manually chosen to be 3 and 0.05, respectively, for the M-distance and Euclidean distance methods. All the integers from 0 to 9 were used for the minimum point count. To quantify the performance, a cost function $c$ for a single trial was devised as:
        \begin{equation}
          c=\frac{1}{n}\sum_{i=1}^n d_{i}+\lambda m. 
          \label{fod_cost}
        \end{equation}  
    Here, 
    \(n\) is the actual number of FODs, \(d_i\) is the Euclidean distance from the \(i\)-th FOD centroid to its closest candidate centroid, \(m\) is the number of candidates that are not associated to any actual FOD, and \(\lambda\) is a positive weighting factor (chosen to be \(0.05\)). 
    
\begin{figure}[thpb]
      \centering
      \framebox{\parbox{1.0\textwidth}{ \includegraphics[width=1.0\textwidth]{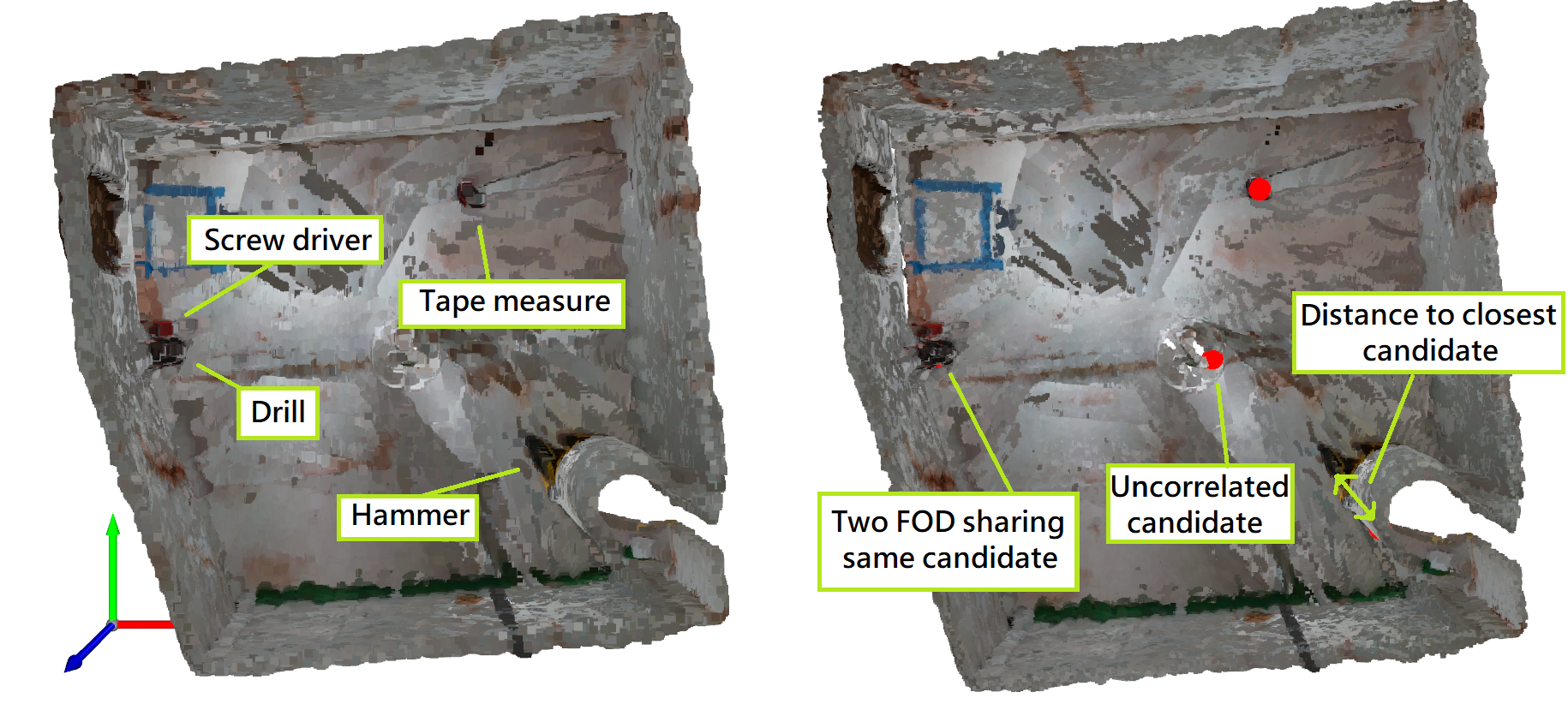}}}
      \caption{A representative example from the physical trials illustrating the associations between the actual and candidate FOD locations. The true FOD locations are shown in the denoised point cloud (left) and the candidates generated using the M-distance method are shown as red dots (right).}
      \label{cost_ex}
   \end{figure}

    Note that multiple FODs can be associated to the same candidate 
    if the distance between them is sufficiently low, as multiple FODs can be captured in the same picture taken from a particular candidate location. A representative example of multiple and missing actual-candidate FOD associations is shown in Figure \ref{cost_ex}. Here, the drill and the screw driver are sufficiently close to share a single candidate without incurring a high distance cost. The candidate at the center of the tank is, however, not associated with any FOD, which results in \(m=1\) for this trial. The closest candidate for the hammer is biased toward the high noise region near the column, which constitutes a large distance cost. 
    
    We compute the mean cost over all the trials for all possible combinations of the parameters, and select the set of parameter values resulting in the minimum cost for the M-distance and Euclidean distance methods separately. The M-distance method achieves a slightly lower cost of (0.493) than the Euclidean metric (0.508). For the M-distance method, the selected threshold, clustering cutoff, and minimum point count values are 2.75, 0.345 m, and 0, respectively. For the Euclidean method, the corresponding values are 0.030 m, 0.279 m, and 4, respectively.
    
  \subsubsection{Evaluation of FOD Candidate Generation Methods}

  The results for FOD candidate generation in a set of 15 test trials are shown in Table \ref{comp_test}. We observe that the total number of unassociated candidate points, prior to clustering, is much smaller for the M-distance method as compared to its Euclidean distance counterpart. 
  In fact, a Wilcoxon signed-rank test on the two test results samples yields a single tailed $p$-value of 0.014. This indicates that the M-distance method has significantly better noise reduction capability at the conventional \(\alpha = 0.05\) significance level.
  
  Correspondingly, on an average, the M-distance method also performs better than the Euclidean distance method in terms of the number of unassociated candidates (after clustering) per trial. This indicates that the former method is expected to be more precise, i.e., avoid false positive during FOD detection. However, both the methods have quite large standard deviations. A direct pairwise comparison shows that the M-distance method yields a smaller number of unassociated candidates in 10 of the 15 trials. It, however, performs particularly badly in 2 of the trials, 
  as discussed below with examples, thereby causing a large standard deviation. The 
  Wilcoxon signed-rank test, with a resultant single tailed $p$-value of 0.176, shows that the overall performance difference is substantial, even though it is not statistically significant. 
  It is useful to mention here that the test is under-powered 
  due to the small sample size and large variance.
 
  The two methods are equivalent with respect to the average distance of the actual FODs from their associated candidate FODs. This is also borne out by the Wilcoxon test that yields a single tailed $p$ value of 0.56. 
  Overall, these results suggest that the M-distance method is likely to reduce the number of false positive FOD detections as compared to the baseline Euclidean method, without causing any appreciable decrease in the quality of the FOD photos. 
  
\begin{table}[h]
\centering
\caption{Performance Comparison over 15 Physical Test Trials with a Total of 54 FODs}
\vspace{2mm}
\resizebox{\columnwidth}{!}{
\begin{tabular}{|c|c|c|c|c|}
\hline
\textbf{}  & \textbf{Total Candidates} & \textbf{Unassociated Points} & 
\textbf{Unassociated Candidates} & \textbf{Distance Error} \\
\textbf{}  &  & \textbf{per Trial} & \textbf{per Trial} & \textbf{per FOD (in $m$)} \\ 
\hline
\textbf{M-Distance} & 72  &  \(127.2\pm 152.9\) & \(2.2\pm 1.56 \)  &   \(0.372 \pm 0.348\)                                    \\
\textbf{L2 Distance} & 78  & \(269.4 \pm 312.5\)  & 2.6 \(\pm\) 1.02&   0.373 \(\pm\) 0.328                                    \\
\hline
\end{tabular}}
      \label{comp_test}
\end{table}

\begin{figure}[thpb]
      \centering
      \framebox{\parbox{1.0\textwidth}{ \includegraphics[width=1.0\textwidth]{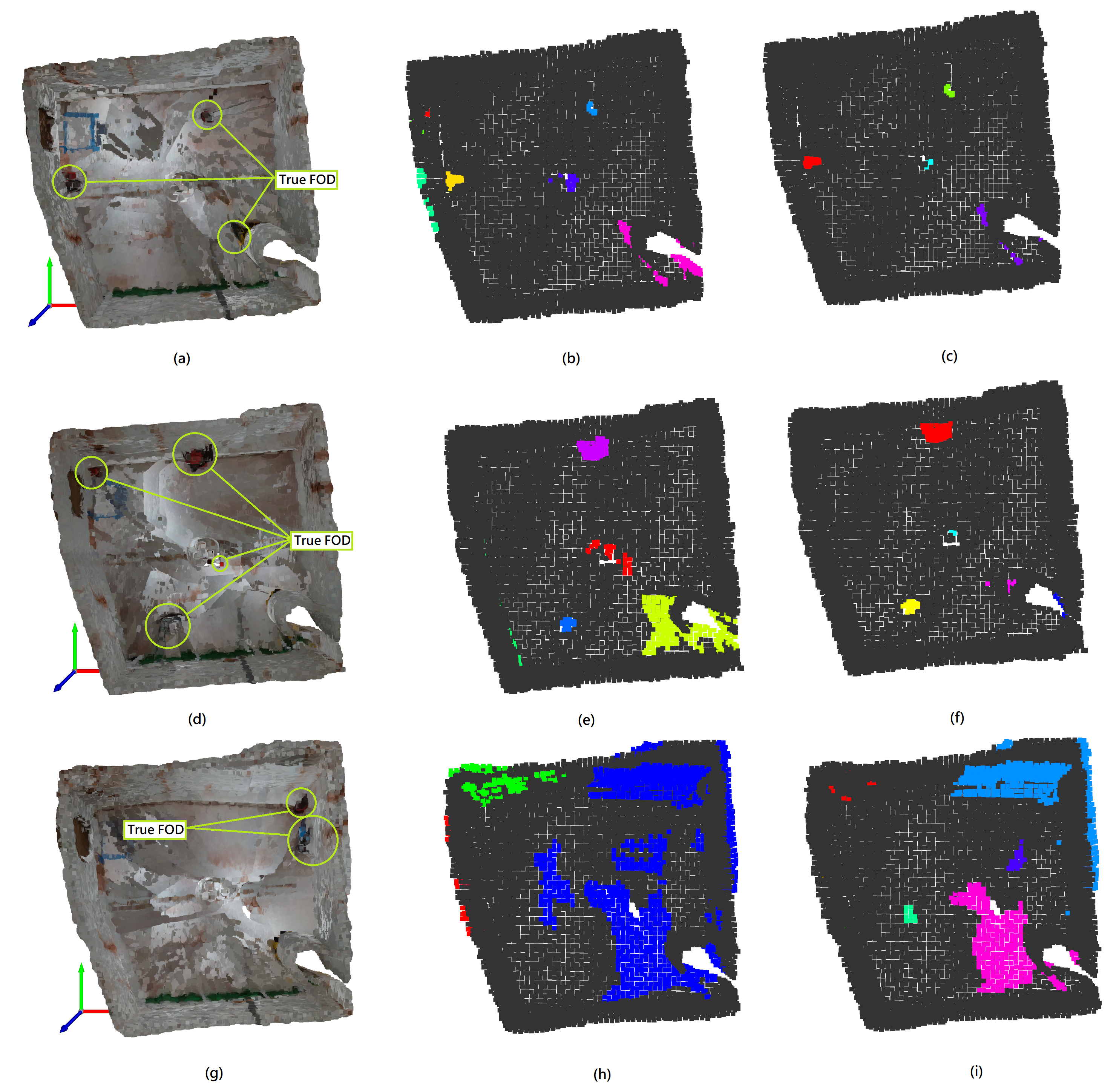}}}
      \caption{Selected examples from physical test trials illustrating the difference in performance between the Euclidean and M-distance methods. Each row represents a single trial. Column 1 shows the denoised point cloud with circled FOD locations; column 2 shows the candidate FOD clusters generated by the Euclidean method, with each cluster highlighted using a different hue; column 3 shows the corresponding clusters generated by the M-distance method. The M-distance method yields substantially less number of candidate points that are not associated with any FOD, and reduces the number of candidate clusters in the first two rows.}
      \label{samples}
   \end{figure}

Figure \ref{samples} shows selected samples from the test set. 
Each row comes from a trial and column 1 shows the point cloud with the true FOD location labeled. Columns 2 and 3 show the candidate clusters generated by the M-distance and Euclidean method, respectively. For all the three trials, the M-distance method shows a lower noise level based on the number of unassociated candidate points. Especially for row 1, it successfully eliminates two noise clusters, one on the left and the other behind the column, such that the candidate centroid is closer to the true FOD location. For the trial in row 2, 
while the column noise is vastly reduced by the M-distance method, the remaining noise is still large enough to form a false cluster. However, it is able to eliminate one noise cluster at the lower left wall that is present in the Euclidean method, while both the methods miss one FOD on the top left corner. Row 3 shows a trial with substantial localization errors due to a misplaced top wall and large noise on the ground. As expected, both the methods perform quite poorly in this case. Although the M-distance method reduces the noise volume, such reduction causes the noisy points to be more disjoint, which inadvertently leads to more candidate clusters than the Euclidean method. This observation explains why the M-distance method has more unassociated candidates than the Euclidean method in a few trials.



 \subsubsection{FOD Detection Performance}
   
The test set was further processed along the FOD identification pipeline, where the cost maps and corresponding waypoints were generated for each trial based on the clusters obtained from the M-distance method. Subsequently, the robot took a photo inside the tank at each location specified by the waypoints to capture the associated FOD candidates. 
 
 The detection rate of the photo set is shown in Table \ref{rate_phy}. Similar to the simulation study, a FOD is considered detected if any portion of the FOD appears in the photo set of a trial. Both the drill and the sander have a perfect recall rate, most likely due to their large heights that cause substantial deviations from the ground plane. 
 On the other hand, the screw driver has the lowest recall, followed by the crimper. The screw driver has the least volume among all the FOD types, which makes it especially challenging to detect. While the crimper has a relatively large volume, it is mostly planar with a height similar to the screw driver when laid on the ground, thereby yielding small deviation values. 
 
   \begin{table}[h]
\centering
  \caption{FOD Detection Rates on Physical Test Trials}
  \vspace{2mm}
\begin{tabular}{|c|c|c|c|}
\hline
\textbf{Type} & \textbf{Detected} & \multicolumn{1}{l|}{\textbf{Missed}} & \multicolumn{1}{l|}{\textbf{Recall}} \\ \hline
Drill & 9 & 0  & 1 \\
Sander & 5  & 0   & 1                                      \\
Screw Driver & 8 & 6 & 0.57                            \\
Hammer  & 7 & 2    & 0.78       \\
Crimper & 5  & 2     & 0.71   \\
Tape Measure & 8     & 2         & 0.8 \\
\hline
Total & 42    & 12  & 0.77 \\
\hline
\end{tabular}
      \label{rate_phy}
\end{table}
 
 The overall recall rate is found to be 0.77. The relevancy of the photo set is also analyzed similar to the simulation study. Out of the 72 photos taken by the robot from the 15 trials, 40 contain one or more FODs and 32 contain no FOD. This constitutes a precision of 0.56. 
 Thus, we note that the precision is higher than that for simulation, while the recall is lower. 
 This happens potentially due to our choice of \(\lambda= 0.05 \) in the grid search cost function, which results in a threshold biased toward reducing the number of irrelevant candidates rather than yielding candidates in close proximity to the true FOD locations. 
   
 The resultant FOD photos were assembled into an online anonymous questionnaire survey. Unlike the simulation study, all the trials are included in a single questionnaire and each participant, therefore, labeled the photos from all the trials. A total of 12 responses were received from undergraduate and graduate students, with an age range from 18-30. 83.3\% of the responders were males and 16.7\% were females. The results are compiled into a confusion matrix and reported in Table \ref{con_mat_id_phy}. 
 
 As expected, the human labeling performance for actual photos taken during physical trials is highly accurate (0.981), and better than that for simulation due to the absence of rendering issues. The most mislabeled item is the tape measure, which has only 2 instances each (out of 102) of being wrongly identified as a drill and screw driver, respectively. A possible cause for this mislabeling is that all these tools have the same black and red colored body. Just 1 photo containing FODs is mislabeled as having no FOD, where the FODs are far away from the robot. Only 2 photos with no FODs (out of 384) are labeled as ``Not Sure'', where both are photos of the loose end of a hose behind the column. These results show that if any FOD is captured in a photo taken by the robot, it is highly likely that a (remotely located) human would be able to detect the FOD just by looking at the corresponding photo.
 


\begin{table}[h]
    \centering
    \caption{Confusion Matrix for Human Labeled FOD Photos Taken over 15 Physical Test Trials ($n=12$) }  
    \vspace{2mm}
    \resizebox{\columnwidth}{!}{
    \begin{tabular}{cc|cccccccc|}
        \cline{3-10}
        &  & \multicolumn{8}{c|}{Actual Type}                                                                             \\ \cline{3-10} 
         & & \multicolumn{1}{c}{\textbf{Drill}} & \textbf{Sander} & \textbf{Screw Driver} & \textbf{Hammer} & \textbf{Crimper} & \textbf{Tape Measure} & \textbf{No FOD} & \textbf{Missed FOD} \\ \hline
        \multicolumn{1}{|c|}{\multirow{8}{*}{\rotatebox[origin=c]{90}{Labeled Type}}} 
                                & \textbf{Drill}        & 141 &	0	&1&	0&	0	&2&	0&0 \\
        \multicolumn{1}{|c|}{}  & \textbf{Sander}       & 0	& 82&	0	&0&	0&	0&	0&	0 \\
        \multicolumn{1}{|c|}{}  & \textbf{Screw Driver} & 1&	0&123&	0&0&	2&0&	0\\
        \multicolumn{1}{|c|}{}  & \textbf{Hammer}       & 1 &	0&0&	108&	0&	0&0&0 \\ 
        \multicolumn{1}{|c|}{}  & \textbf{Crimper}      & 0&2&0&	0&	68&	0&0&	0\\
        \multicolumn{1}{|c|}{}  & \textbf{Tape Measure} & 0 & 0 &	1&0&	0&	98&	0&0 \\
        \multicolumn{1}{|c|}{}  & \textbf{No FOD}       & 0 &	0&	0&	0&	0&	0&	382&1 \\
        \multicolumn{1}{|c|}{}  & \textbf{Not Sure}     & 0&0&0&	0&0&	0&2&	0\\ 
        \hline
    \end{tabular}
    }
    \label{con_mat_id_phy}
\end{table}  

\section{Discussion}
\label{discussion}
There are a few limitations of our approach, as seen in the experimental results. The primary assumption of using M-distance is that the deviation from the reference model is Gaussian. 
However, there are multiple sources of non-Gaussian noise during the SLAM process. One source is localization uncertainty, which causes feature duplication at the erroneous robot location. This poses problem when using the M-distance as it treats the entire duplicate feature as deviation and causes false positives. Another problematic noise type is associated with the depth camera, which occurs around the edge of an object where the depth gradient is very high. Normally, the edge noises are removed during the SLAM process when the robot travels behind the obstacles. However, this does not happen for confined space structures such as the I-beam on the wall. These noise sources can be potentially dealt with by transitioning to non-Gaussian learning methods such as mixture models.

The clustering method also introduces an error in conjunction with this duplicated feature problem. When there is a large patch of false positive points, the hierarchical clustering algorithm sometimes clusters the true FOD points with the false positive points into one group. As a result, the center of the cluster mass meanders more toward the false positive, and sometimes causes the true positive to disappear from the camera's field of view. A work-around for this issue could be including the cluster size information during waypoint generation. Another method would be to include color information in the feature vector and perform a point cloud version of color image segmentation.

The training sample collection process can be tedious. If the mean point cloud is chosen as the reference, then the training sample is readily available and covariance fitting does not result in any extra work. However, if an accurate CAD model is used,
collecting the training set only for covariance fitting may not be cost-efficient. This issue can be addressed by moving the learning process online, where a CAD model with Euclidean distance is used as the initial method. After each inspection session, a operator labels and removes all the FOD points, and the remaining points are used as the training samples, where the points in the CAD model are updated by the training points in a Gaussian mixture fashion.

Last but not the least, the trade-off between high FOD recall and photo precision is tricky, since the parameter search cost function does not directly consider the waypoint generation process and camera specifications. While the cost function provides a unified way of comparing different discrepancy metrics, hand tuning might be more advantageous for achieving a specific precision-recall rate.

\section{Conclusions}

In this paper, we propose a remote human-assisted, visual mapping-based probabilistic FOD detection system for confined spaces within large marine vessels. A generic water tank, with a publicly released CAD model, is used as the representative confined space for system development and testing. A local Mahalanobis distance-driven outlier identification method forms the core of our system, which enables identification of candidate FODs by quantifying the discrepancies between the offline FOD-less maps (or, the tank CAD model) and the online maps acquired by a mobile ground robot. Camera photos taken by the robot from the candidate FOD locations are then provided to humans for final labeling, which results in a high detection accuracy.
An initial simulation study, followed by extensive physical trials on a scaled-down tank prototype, demonstrate the effectiveness of our detection system. 

In the future, we would like to modify our system for other kinds of inspection tasks, such as identification and monitoring of tank defects and damages. We would also like to explore the potential benefits of multi-robot coordination in inspecting large confined spaces efficiently. To this end, we plan to integrate our detection system with novel locomotion capabilities to facilitate precise manipulation in tight spaces; on tall, vertical structures; and, for a variety of repair tasks. We are also interested in adapting the inspection system for underwater environments (when the spaces are filled with water), and investigating alternate sensing, odometry, and mapping techniques for such environments.  

\section*{Acknowledgment}
We gratefully acknowledge John Stewart for his help in building the water tank prototype and Prof. Santosh Devasia for many useful discussions. This work was supported by the Naval Engineering Education Consortium (NEEC) award number N00174-20-1-0003. Any opinions,  findings,  and conclusions  or recommendations  expressed  in  this paper are  those  of  the authors and do not necessarily reflect the views of the US Navy.

 \bibliographystyle{elsarticle-num} 
 \bibliography{references}

\begin{thebibliography}{10}
\expandafter\ifx\csname url\endcsname\relax
  \def\url#1{\texttt{#1}}\fi
\expandafter\ifx\csname urlprefix\endcsname\relax\def\urlprefix{URL }\fi
\expandafter\ifx\csname href\endcsname\relax
  \def\href#1#2{#2} \def\path#1{#1}\fi

\bibitem{navy}
CBO,
  \href{https://www.cbo.gov/system/files/2019-10/55685-CBO-Navys-FY20-shipbuilding-plan.pdf}{{US}
  {N}avy maintenance cost} (2020).
\newline\urlprefix\url{https://www.cbo.gov/system/files/2019-10/55685-CBO-Navys-FY20-shipbuilding-plan.pdf}

\bibitem{Bandyopadhyay2018}
T.~Bandyopadhyay, R.~Steindl, F.~Talbot, N.~Kottege, R.~Dungavell, B.~Wood,
  J.~Barker, K.~Hoehn, A.~Elfes, Magneto: A versatile multi-limbed inspection
  robot, in: IEEE/RSJ Int. Conf. Intell. Robot. Syst., 2018, pp. 2253--2260.

\bibitem{kakogawa2020}
A.~Kakogawa, S.~Ma, A multi-link in-pipe inspection robot composed of active
  and passive compliant joints, in: IEEE/RSJ Int. Conf. Intell. Robot. Syst.,
  2020, pp. 6472--6478.

\bibitem{Virgala2020}
I.~Virgala, M.~Kelemen, P.~Božek, Z.~Bobovský, M.~Hagara, E.~Prada,
  P.~Oščádal, M.~Varga, Investigation of snake robot locomotion
  possibilities in a pipe, Symmetry 12~(6) (2020).

\bibitem{versatrax}
E.~Technologies, \href{https://eddyfi.com/en/application/pipelines}{Versatrax
  pipe inspection crawler} (2021).
\newline\urlprefix\url{https://eddyfi.com/en/application/pipelines}

\bibitem{Owan2017}
P.~Owan, J.~Garbini, S.~Devasia, Addressing agent disagreement in
  mixed-initiative traded control for confined-space manufacturing, in: IEEE
  Int. Conf. Adv. Intell. Mechatronics, 2017, pp. 227--234.

\bibitem{Owan2020}
P.~Owan, J.~Garbini, S.~Devasia, Faster confined space manufacturing
  teleoperation through dynamic autonomy with task dynamics imitation learning,
  IEEE Robot. Autom. Lett. 5~(2) (2020) 2357--2364.

\bibitem{Han2020}
Z.~Han, J.~Allspaw, G.~LeMasurier, J.~Parrillo, D.~Giger, S.~R. Ahmadzadeh,
  H.~A. Yanco, Towards mobile multi-task manipulation in a confined and
  integrated environment with irregular objects, in: IEEE Int. Conf. Robot.
  Autom., 2020, pp. 11025--11031.

\bibitem{Tripicchio2018}
P.~Tripicchio, M.~Satler, M.~Unetti, C.~A. Avizzano, Confined spaces industrial
  inspection with micro aerial vehicles and laser range finder localization,
  Int. J. Micro Air Veh. 10~(2) (2018) 207--224.

\bibitem{Preston2018}
V.~Preston, T.~Salumäe, M.~Kruusmaa, Underwater confined space mapping by
  resource-constrained autonomous vehicle, J. Field Robot. 35~(7) (2018)
  1122--1148.

\bibitem{Brogaard2020}
R.~Y. Brogaard, M.~Zajaczkowski, L.~Kovac, O.~Ravn, E.~Boukas, Towards
  {UAV}-based absolute hierarchical localization in confined spaces, in: IEEE
  Int. Symp. Safety Security Rescue Robot., 2020, pp. 182--188.

\bibitem{Akbari2020}
A.~Akbari, P.~S. Chhabra, U.~Bhandari, S.~Bernardini, Intelligent exploration
  and autonomous navigation in confined spaces, in: IEEE/RSJ Int. Conf. Intell.
  Robots Syst., 2020, pp. 2157--2164.

\bibitem{DePetris2020}
P.~De~Petris, H.~Nguyen, T.~Dang, F.~Mascarich, K.~Alexis, Collision-tolerant
  autonomous navigation through manhole-sized confined environments, in: IEEE
  Int. Sym. Safety Security Rescue Robot., 2020, pp. 84--89.

\bibitem{Azpurua2021}
H.~Azpúrua, A.~Rezende, G.~Potje, et~al., Towards semi-autonomous robotic
  inspection and mapping in confined spaces with the {E}speleo{R}obô, J.
  Intell. Robot. Syst. 101~(4) (2021) 69.

\bibitem{Ozturk2016}
S.~Öztürk, A.~E. Kuzucuoğlu, A multi-robot coordination approach for
  autonomous runway foreign object debris ({FOD}) clearance, Robot. Auton.
  Syst. 75 (2016) 244--259.

\bibitem{Cao2018}
X.~Cao, P.~Wang, C.~Meng, X.~Bai, G.~Gong, M.~Liu, J.~Qi, Region based {CNN}
  for foreign object debris detection on airfield pavement, Sensors 18~(3)
  (2018) 737.

\bibitem{Gao2021}
Q.~Gao, R.~Hong, Y.~Chen, J.~Lei, Research on foreign object debris detection
  in airport runway based on semantic segmentation, in: 2nd Int. Conf. Comput.
  Data Sci., 2021, pp. 1--3.

\bibitem{Jing2022}
Y.~Jing, H.~Zheng, C.~Lin, W.~Zheng, K.~Dong, X.~Li, Foreign object debris
  detection for optical imaging sensors based on random forest, Sensors 22~(7)
  (2022).

\bibitem{Lai2020}
Y.-K. Lai, {Foreign object debris detection method based on fractional
  {F}ourier transform for millimeter-wave radar}, J. Appl. Remote Sensing
  14~(1) (2020) 1 -- 15.

\bibitem{Ni2020}
P.~Ni, C.~Miao, H.~Tang, M.~Jiang, W.~Wu, Small foreign object debris detection
  for millimeter-wave radar based on power spectrum features, Sensors 20~(8)
  (2020) 2316.

\bibitem{Futatsumori2021}
S.~Futatsumori, N.~Yonemoto, N.~Shibagaki, Y.~Sato, K.~Kashima, Detection
  probability estimation of 96 {GH}z millimeter-wave airport foreign object
  debris detection radar using measured radar cross section characteristics,
  in: Eur. Conf. Antennas Propag., 2021, pp. 1--4.

\bibitem{Fizza2021}
G.~Fizza, S.~M. Idrus, F.~Iqbal, W.~H.~W. Hassan, N.~Shibagaki, K.~Kashima,
  A.~Hamzah, S.~Ambran, T.~Kawanishi, Line of sight visibility analysis for
  foreign object debris detection system, J. Phy.: Conf. Series 1878~(1) (2021)
  012006.

\bibitem{Liu2021}
T.~Liu, H.~Cui, Y.~Wang, M.~Zhai, J.~Zhang, Z.~Wei, Adaptive leakage
  cancelation method in frequency modulated continuous wave radar for foreign
  object debris detection, Int. J. RF Microw. C E 31~(3) (2021) e22546.

\bibitem{Zhong2021}
J.~Zhong, X.~Gou, Q.~Shu, X.~Liu, Q.~Zeng, A {FOD} detection approach on
  millimeter-wave radar sensors based on optimal {VMD} and {SVDD}, Sensors
  21~(3) (2021) 997.

\bibitem{Kniaz2014}
V.~V. {Kniaz}, {A Fast Recognition Algorithm for Detection of Foreign 3{D}
  Objects on a Runway}, ISPRS - Int. Archives Photogrammetry, Remote Sensing
  Spatial Inf. Sci. XL3 (2014) 151--156.

\bibitem{Mund2015}
J.~Mund, A.~Zouhar, L.~Meyer, H.~Fricke, C.~Rother, Performance evaluation of
  {L}i{DAR} point clouds towards automated {FOD} detection on airport aprons,
  in: Proc. Int. Conf. Appl. Theory Autom. Command Control Syst., 2015, pp.
  85--94.

\bibitem{Elrayes2019}
A.~Elrayes, M.~H. Ali, A.~Zakaria, M.~H. Ismail, Smart airport foreign object
  debris detection rover using {L}i{DAR} technology, Internet Things 5 (2019)
  1--11.

\bibitem{Xu2018}
H.~Xu, Z.~Han, S.~Feng, H.~Zhou, Y.~Fang, Foreign object debris material
  recognition based on convolutional neural networks, {EURASIP} J. Image Video
  Proc. 21 (2018) 1--10.

\bibitem{Zhang2019}
W.~Zhang, X.~Liu, J.~Yuan, L.~Xu, H.~Sun, J.~Zhou, X.~Liu, {RCNN}-based foreign
  object detection for securing power transmission lines ({RCNN4SPTL}),
  Procedia Comput. Sci. 147 (2019) 331--337.

\bibitem{Haotian2021}
S.~Haotian, L.~Tong, W.~Pu, X.~Liang, Z.~Hongwei, Foreign object detection of
  electric transmission line with dynamic federated learning, {IOP} Conf. Ser.:
  Earth Environ. Sci. 791~(1) (2021) 012159.

\bibitem{Kuo2022}
R.~J. Kuo, F.~F. Nursyahid, Foreign objects detection using deep learning
  techniques for graphic card assembly line, J. Intell. Manuf. (2022).

\bibitem{Xiong2020}
H.~Xiong, J.~Wu, Q.~Liu, Y.~Cai, Research on abnormal object detection in
  specific region based on {M}ask {R-CNN}, Int. J. Adv. Robot. Syst. 17~(3)
  (2020) 1729881420925287.

\bibitem{Latimer2019}
K.~Latimer, Remote visualization and detection of foreign object debris in
  aerospace manufacturing using a low-cost depth camera, Master's thesis,
  University of Washington (2019).

\bibitem{Kahn2013}
S.~Kahn, U.~Bockholt, A.~Kuijper, D.~W. Fellner, Towards precise real-time 3{D}
  difference detection for industrial applications, Comput. Ind. 64~(9) (2013)
  1115--1128.

\bibitem{rtabmap}
M.~Labbé, F.~Michaud, {RTAB-M}ap as an open-source lidar and visual
  simultaneous localization and mapping library for large-scale and long-term
  online operation, J. Field Robot. 36~(2) (2019) 416--446.

\bibitem{cloudcompare}
EDF,
  \href{https://www.cloudcompare.org/doc/wiki/index.php?title=Local\_Statistical\_Test}{Cloud{C}ompare
  local statistical test} (2020).
\newline\urlprefix\url{https://www.cloudcompare.org/doc/wiki/index.php?title=Local\_Statistical\_Test}

\bibitem{reliability_weight}
M.~Galassi, J.~Davies, J.~Theiler, B.~Gough, G.~Jungman, M.~Booth, F.~Rossi,
  \href{https://www.gnu.org/software/gsl/doc/html/statistics.html#c.gsl\_stats\_wvariance}{Gnu
  scientific library- reference manual} (2021).
\newline\urlprefix\url{https://www.gnu.org/software/gsl/doc/html/statistics.html#c.gsl\_stats\_wvariance}

\bibitem{costmap}
E.~Marder-Eppstein, D.~V. Lu, D.~Hershberger, costmap\_2d,
  \url{http://wiki.ros.org/costmap_2d}, accessed: 2022-07-28.

\end{thebibliography}
 \pagebreak
 
\appendix
\section{Modified Rtap-Map Configuration}
\label{rtabmap_param}
\subsection{Global Parameters}
\begin{itemize}
\item   frame\_id=base\_footprint 
\item   odom\_frame\_id=odom
\item   approx\_rgbd\_sync=false
\item   odom\_tf\_angular\_variance=1    
\item   odom\_tf\_linear\_variance=1
\end{itemize}
\subsection{Nodes Parameters}
\begin{itemize}
 \item Node Name: map\_assembler
    \begin{itemize}
      \item  cloud\_output\_voxelized=false
      \item  Grid/RangeMax=0.5
      \item  Grid/cloud\_subtrack\_filtering\_min\_neighbors=10
      \item  Grid/DepthDecimation=1
    \end{itemize}
    
      \item  Node Name: map\_optimizer
          \begin{itemize}

  \item publish\_tf=false
  \item odom\_frame\_id=\$(arg odom\_frame\_id)
    \end{itemize}

    \item  Node Name: rtabmap
        \begin{itemize}
        \item grid\_map=/map

      \item Grid/FromDepth=true
      \item Grid/CellSize=0.01
      \item Grid/DepthDecimation=1
      \item Grid/MaxGroundHeight=0.05
      \item Grid/MaxObstacleHeight=1
    \end{itemize}
  \end{itemize}
\end{document}